# Epistocracy Algorithm: A Novel Hyper-heuristic Optimization Strategy for Solving Complex Optimization Problems

Seyed Ziae Mousavi Mojab[1], Seyedmohammad Shams[2], Hamid Soltanian-Zadeh[3], Farshad Fotouhi[4]

[1,4] Dept. of Computer Science, Wayne State University, Detroit, MI 48202, USA
[2,3] Dept. of Radiology, Henry Ford Health System, Detroit, MI 48202, USA

**Abstract.** This paper proposes a novel evolutionary algorithm called Epistocracy which incorporates human socio-political behavior and intelligence to solve complex optimization problems. The inspiration of the Epistocracy algorithm originates from a political regime where educated people have more voting power than the uneducated or less educated. The algorithm is a self-adaptive, and multi-population optimizer in which the evolution process takes place in parallel for many populations led by a council of leaders. To avoid stagnation in poor local optima and to prevent a premature convergence, the algorithm employs multiple mechanisms such as dynamic and adaptive leadership based on gravitational force, dynamic population allocation and diversification, variance-based step-size determination, and regression-based leadership adjustment. The algorithm uses a stratified sampling method called Latin Hypercube Sampling (LHS) to distribute the initial population more evenly for exploration of the search space and exploitation of the accumulated knowledge. To investigate the performance and evaluate the reliability of the algorithm, we have used a set of multimodal benchmark functions, and then applied the algorithm to the MNIST dataset to further verify the accuracy, scalability, and robustness of the algorithm. Experimental results show that the Epistocracy algorithm outperforms the tested state-of-the-art evolutionary and swarm intelligence algorithms in terms of performance, precision, and convergence.

## 1 Introduction

Evolutionary computation (EC) is a subfield of artificial intelligence that encompasses methods mimicking mechanisms of biological evolution to solve various optimization problems. An optimization problem essentially requires finding a set of parameters $\vec{x} = (x_1, \ldots, x_n) \in S$ of the current system, such that a certain quantity $f: S \to \mathbb{R}$ is maximized (or minimized) $\forall \vec{x} \in S : f(\vec{x}) \leq f(\vec{x}^*)$.

Over the past few decades, many state-of-the-art evolutionary algorithms such as Genetic algorithm (GA) and Evolutionary Strategies (ES) have been proposed for applications where a well-defined or closed-form solution does not exist [1]. Genetic algorithm was developed by John Holland in the early 1970s [2]-[4] mimicking Darwinian theory of survival of the fittest and Evolutionary Strategies founded by Rechenberg and Schwefel in 1965 [5]-[7] based on the hypothesis that small mutations occur more commonly than large mutations. Both Genetic algorithm and Evolutionary Strategies rely on the concept of population, representing potential solutions to the optimization problem which iteratively undergo genetic operators to improve their fitness score. While Genetic algorithms use a binary string of digits to represent solutions and use both mutation and recombination as genetic operators, in Evolutionary Strategies a fixed-length real-valued vector is used for representation, and only mutation is used as a primary search operator. In evolutionary algorithms, the recombination operator performs an information exchange, and the mutation operator generates variations of the solutions and increases the diversity among the population. The selection operator, however, makes better individuals to survive and reproduce.

Another subset of nature-inspired algorithms is Swarm Intelligence (SI) which is based on collective behavior of a decentralized, self-organizing network of agents such as bird flocks or honeybees. In SI algorithms, multiple agents can locally interact and exchange heuristic information which leads to the emergence of global behavior of adaptive search and optimization. Particle Swarm Optimization (PSO) is an example of swarm intelligence proposed by Eberhart and Kennedy in 1995 [8]. This algorithm is inspired by social behavior of bird flocking and fish schooling. Similar to GA, PSO is initialized with a population of random candidate solutions that are improved iteratively over time, however, unlike GA has no evolution operators such as recombination and mutation. Despite the fact that PSO is a powerful and effective optimization technique, it still suffers from stagnation and premature convergence [9], [10]. Several solutions including inertia weight, and time-varying coefficients have been proposed to eliminate these problems [11], [12].

The Artificial Bee Colony (ABC) is another popular swarm intelligence-based algorithm which is inspired by the foraging behavior of the honeybees. ABC consists of three groups of bees: employed bees, onlookers, and scouts that have different roles in the optimization process. ABC is simple, easy to implement, and highly flexible [13]. This algorithm was first proposed by Dervis Karaboga in 2005 [14] to optimize numerical problems. Since then, many variants of ABC have been introduced to increase the population diversity and avoid premature convergence [15], [16].

Cuckoo Search Algorithm (CSA) is one of the latest swarm intelligence-based algorithms developed by Yang and Deb in 2009 [17]. This algorithm is inspired by natural behavior of cuckoos who lay their eggs in other birds' nests for breeding. Compared to other approaches, Cuckoo requires fewer numbers of parameters to be fine-tuned. In 2018, Mareli *et al*. [18] developed three new Cuckoo search algorithms using linear, exponential and power increasing switching parameters to maintain an optimum balance between local and global exploration and increase the efficiency of CS algorithm. In 2019, Li *et al*. [19] proposed a new variant of CSA called I-PKL-CS algorithm which employs self-adaptive knowledge learning strategies to mitigate premature convergence and poor balance between exploitation and exploration. I-PKL-CS exploits individual and population knowledge learning to improve the quality of solutions and convergence rate.

There exist many real-world applications for EC. In [20], the genetic algorithm was used to decrease the dimension of the data and to optimize the weights and biases of the neural network in ECG signal classification. Xi *et al.* [21] used PSO to improve the performance of their neural network in order to assess the hazard of earthquake-induced landslide. Kim *et al.* [22] used self-adaptive Evolutionary Strategies to optimize the parameters of an autonomous car controller. Prakash *et al.* [23] employed the Cuckoo Search algorithm to perform job scheduling and resource allocation on the grid. Yeh *et al.* [24] used ABC to optimize a bee recurrent neural network to generate a novel approximate model for predicting network reliability.

The selection of an evolutionary approach can drastically reduce the amount of time needed for finding an optimal solution. According to several studies, evolutionary algorithms, in general, suffer from various problems such as limited searching ability [25]-[27], curse of dimensionality and scalability [28], [29], premature convergence and stagnation [30]-[33], and poor performance which usually occur in the absence of population diversity and adaptability [34]-[36], and due to unbalanced exploration-exploitation capacities [37], [38].

The work reported in this paper was motivated by the fact that optimization algorithms require new explorative and exploitative capabilities along with a dynamic resource allocation technique and diversification strategies to help them converge to the optimal solution at the early stages of the optimization process. There is a need for a new generation of evolutionary algorithms that can avoid entrapment in local optima and prevent premature convergence [39], [30]. To find the optimal solution, these algorithms must employ a directed and goal-oriented search rather than a purely random and stochastic one.

In this paper, we propose a new hyper-heuristic algorithm based on a political regime called Epistocracy where educated people have more voting power (weight) than the uneducated or less educated. The Epistocracy algorithm splits the population into Governors and Citizens based on the performance of the individuals. The Citizens are assigned Governors based on the degree of similarity and the exercise of free will. Once a Citizen is assigned a Governor, they move towards their Governor in an attempt to mimic some of the traits which made their Governor successful. Governors will also try to improve themselves and lead their population to collaboratively search for the optimal solution.

The Epistocracy algorithm is a self-adaptive, and multi-population optimizer in which the evolution process takes place in parallel for many populations led by a council of leaders. To avoid entrapment in poor local optima and to prevent a premature convergence, the algorithm employs multiple mechanisms such as dynamic and adaptive leadership based on gravitational force, dynamic population allocation and diversification, variance-based step-size determination, and regression-based leadership adjustment. The algorithm uses a stratified sampling method called Latin

Hypercube Sampling (LHS) to distribute the initial population more evenly for exploration of the search space and exploitation of the accumulated knowledge.

The rest of the paper is organized as follows. Section 2 describes the overall structure of the Epistocracy algorithm in detail. Experimental results and comparative studies on benchmark test functions along with Convolutional Neural Networks (CNNs) parameter optimization are presented in Section 3. Finally, conclusions and directions for future research are presented in Section 4.

## 2    Epistocracy Algorithm

### 2.1    Overview

The term Epistocracy is derived from the Greek word epistêmê meaning knowledge, knowing, and understanding. John Stuart Mill (1806-1873), the British philosopher and political economist in his book "Mill on Bentham and Coleridge" proposed to give more votes to the better educated [40]. Jason Brennan believes that more competent or knowledgeable citizens must have slightly more political power than less competent citizens [41]. In fact, the problem with democracy is the elimination of the epistemic dimension of democracy. While Democracy is more about the input aspect of the decision-making process, Epistocracy is concerned about the output.

Epistocracy algorithm is multi-population optimization algorithm which seeks to minimize the time taken to find an optimal value for the problem being solved. As an adaptive, hyper-heuristic algorithm, Epistocracy employs problem-related knowledge, and globally aggregated statistics to automatically adjust itself during each run and search through a space of meta-heuristics to find the optimal solution. Epistocracy attempts to incorporate human sociopolitical behavior and intelligence to improve the performance and convergence speed and reduce the probability of getting trapped in local optima compared to other meta-heuristic algorithms.

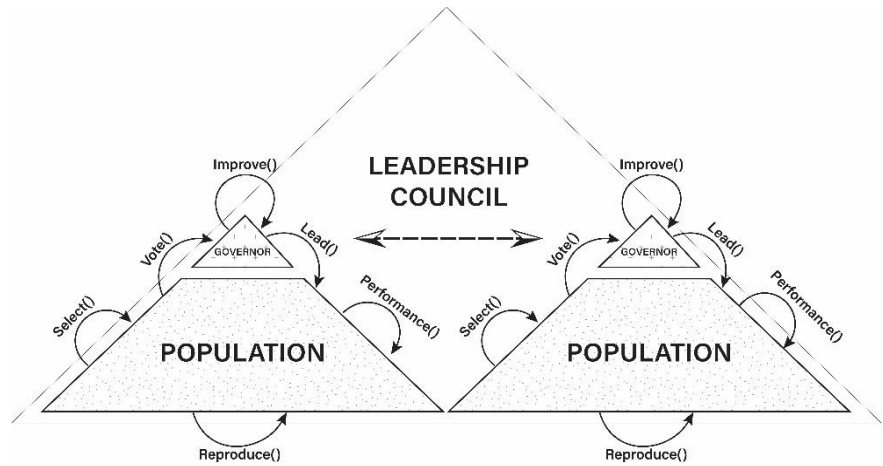

**Fig. 1.** Flow diagram of Epistocracy algorithm.

As illustrated in Fig. 1, the Epistocracy algorithm is made of two primary components: Governors and Citizens. Citizens are individual solutions that are randomly and uniformly created. In each generation, all individuals are evaluated with a pre-defined fitness function. The top-performing individuals (Governors) are then selected through the Select() function to lead the population. Governors are, in fact, a network of cooperative leaders who influence and evolve the generation of the new population via Lead() function. While Governors continuously improve themselves, citizens can vote for governors and affect their position in the government. Information is systematically propagated among citizens and governors. Fig. 2 shows the flowchart of the proposed algorithm.

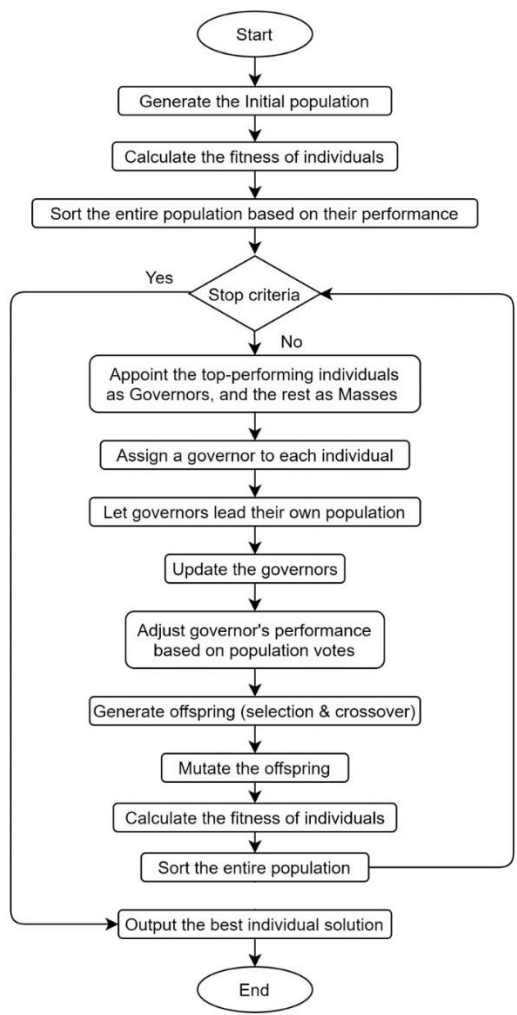

**Fig. 2.** Flowchart of the Epistocracy algorithm with all steps involved from the population generation until outputting the optimal solution.

### 2.2 Generating the Initial Population

The Epistocracy algorithm starts the optimization process by generating a population of random solutions, using a stratified sampling method called Latin Hypercube Sampling (LHS) which was originally proposed by McKay in 1979 [42]. Each individual solution has a set of genes or attributes known as chromosome which are defined using their corresponding upper and lower bounds in the search space. In this algorithm, the set of attributes represent the initial position and level of political knowledge of each individual in the society.

### 2.3 Performance Evaluation

The performance of an individual in the population is evaluated using a pre-defined fitness function. Given the individual's current chromosome, the actual performance is calculated and stored as an individual's "actual performance." The previous actual performance is also recorded for future reference.

Individual solutions are then ranked based on their actual performance (fitness score). The adjusted performance is calculated based on the actual performance of each individual solution. The calculation steps will be explained in detail in the following sections.

### 2.4 Population Separation

Different people demonstrate different understandings of patterns of change and achieve different levels of success and result upon the social hierarchy. This algorithm plans to separate the Governors from Citizens based on the level of success an individual can achieve. The top performers in the population will be considered "Governors" and the rest will be considered "Citizens".

### 2.5 Governor Assignment

Before evolving each individual and moving them around, about five percent of the top-performing individuals in each generation are selected as a set of governors to lead the population and help them improve their performance. Transcending traditional societies, in Epistocracy, governments have no obvious borders, and individuals can follow or vote for any governor anywhere expressing the idea of "Global Village". In the Epistocracy algorithm, each individual is assigned to a governor based on their phenotypic characteristics, and the degree of influence and impact of the governor on the citizen. To that end, the Gravitational Force (1) is used to calculate the magnitude of attraction between each citizen and every governor. A governor with a larger gravitational force has a higher probability to attract a citizen and form a larger territory. However, some citizens may act as rebels and resist against the orders of the befitting authorities and may follow different governors.

$$F = G \times \left(\frac{m_1 \times m_2}{r^2}\right) \qquad (1)$$

In the above equation of the Gravitational Force, $G$ is a constant, and $m_1$, and $m_2$, are the adjusted performances of the governor and citizen respectively. All performances are normalized using the following formula:

$$P_{norm\ i} = \left[\frac{P_i - min(P_{governors})}{max(P_{governors}) - min(P_{governors})}\right]^{-1} \quad (2)$$

In (2), $P_i$ is the individual's actual performance, where $P_{governors}$ is the list of governors' performances.

The Euclidean distance (3) is used to calculate the distance $r$ between a governor and a citizen.

$$dist(x_i, x_j) = \|x_i - x_j\| = \sqrt{\sum_{k=1}^{n}(x_{ik} - x_{jk})^2}$$

$$n = number\ of\ variables \quad (3)$$

To imitate the rebelliousness of citizens, a roulette wheel with the governors' calculated gravitational forces is used to give citizens a freedom of selecting other governors with even a greater dissimilarity (distance). This will help the algorithm to explore the interspace between the governors by moving a citizen across the governments. The selection probability is defined using the following equation:

$$P_j = \frac{G(S_j)}{\sum_{i=1}^{n} G(S_i)} \quad (4)$$

In (4), $n$ is the number of governors. $G$ is the gravitational force of solution $S_j$.

In the next generation, if the assigned governor is overthrown or resigned due to their poor performance or their own population votes, the surviving citizen will choose a new governor from the updated list of governors. If a governor performs poorly, eventually, he will be degraded and may lose all his population and get removed from the current list of governors. This happens when a population's total performance over a certain period of time (an iteration) compared to other populations is very small. In this case, the governor will lose his popularity regardless of his own performance at the time of being selected. In fact, a governor's popularity rests on his credibility and competence, and his performance in leading his population and improving their lives. By adjusting the actual performance of the governor, the governor's rank in the governors list will change. Given that, this governor will have a lower chance to be selected by new citizens who do not have any governor yet.

### 2.6 Leading the Population

In the next step, the Epistocracy algorithm allows governors to lead their own population. Each citizen will take a step of variable length (5) toward his governor to improve his performance and become similar and even better than their governor. The step size is proportional to the distance between the governor and citizen and inversely proportional to the self-improvement of the citizen under the rule of the governor. The following formula is used to calculate the next step of each citizen:

$$S_i = \left(\frac{I_{avg}}{I_{min}}\right) \times \sigma^2 \times d_{i,g} \times \varphi \tag{5}$$

where $S_i$ is the individual's new step size, and $I_{avg}$ is the average improvement of the governor's sub-population (7). $I_{min}$ is the minimum improvement in the population. $\sigma^2$ is the variance of the sub-population, and $d_{i,g}$ is the Euclidean distance between the individual and its designated governor. $\varphi$ is the rate of change equal to 0.1. The self-improvement is calculated as follows:

$$I_i = (P_{old\ i} - P_{actual\ i}) \tag{6}$$

The self-improvement is the difference between the old and the current actual performance of the citizen. The average improvement is then calculated by:

$$I_{avg} = \frac{1}{n} \times \sum_{i=1}^{n} I_i \tag{7}$$

In (7), $n$ is the size of the governor's sub-population. The average improvement is an important factor for the step size determination. To avoid missing any minima or maxima, a smaller step will be taken when a larger improvement is achieved, and a larger step will be taken when a smaller improvement is obtained.
The population variance is given by the following formula:

$$\sigma^2 = \frac{\sum(x_i - \mu)^2}{n} \tag{8}$$

To reflect the diversity of the society, if a citizen by taking a new step becomes exactly similar to his governor or another citizen in the same population, the citizen will be mutated to save the system resources. This also helps the algorithm to avoid division by zero in calculating the gravitational force when the distance between a citizen and his governor becomes zero.

### 2.7 Improving Governors

Similar to citizens, governors will also improve themselves by taking a step in a direction that hopefully increases their performance. To that end, the variance of governors' population is calculated, helping governors converge toward a location with the highest possibility of finding the optimal solution. The next step size of the governor is calculated like that of a citizen. However, instead of calculating the distance between the governor and citizen, this time the governor's previous step is considered according to the following formula:

$$S_j = \left(\frac{I_{avg}}{I_j}\right) \times \sigma^2 \times S_{prev\ j} \tag{9}$$

where "previous step" $S_{prev}$ is initialized as:

$$prevStep = (upper_{limit} - lower_{limit}) \times space_{resolution} \tag{10}$$

In (10), $upper_{limit}$ and $lower_{limit}$ are the boundaries of the search space, and the space resolution is initially set to 0.001.

Since the governor is in charge of leading his population and pushing them to the right direction, the algorithm will let the governor take a step only if that step improves his overall performance, otherwise, the governor will stay in his previous place without making any movement.

Since for computing the new step the variance of all governors is used, in the next iterations for the same governor the step size might be different and might help the governor to get improved and consequently positively contribute to the improvement of his population. The following piecewise function (11) shows the conditional step that must be taken by each governor, provided that the step improves the governor's actual performance:

$$S_j = \begin{cases} \left(\frac{I_{avg}}{I_j}\right) \times \sigma^2 \times S_{prev\ j} & if\ \Delta P \leq 0 \\ 0 & otherwise \end{cases} \tag{11}$$

where $\Delta P = P_{actual\ new} - P_{actual\ old}$. This formula is designed for a minimization optimization problem.

## 2.8 Governor's Performance Adjustment

When a population performs well or poorly under a leadership of a governor, the algorithm will adjust the governor's actual performance to allocate the right amount of resources (individuals) to the governor. For example, if a population is performing well, that must increase the trust of the population in the governor. In this case, generally more individuals will be following the governor to help him accomplish the task of finding the optimal solution. If a governor is performing poorly, the governor's actual performance will be lowered accordingly, and eventually, some individuals will leave the governor and follow another governor to improve their quality of life. In other words, like the Epistocratic societies, when a population is under-performing, this will eventually affect the popularity and credibility of the governor. Those people who initially voted for that governor, will shift away from the governor, and try to choose another governor. In each iteration, the population will vote on the performance of the governor, however, these votes have different weights.

The Epistocracy algorithm will compute the average improvement per each population, giving higher weights to individuals who are closer to the governor (and more educated) and lower weights to citizens who are farther away (and less educated) before using the following formula:

$$I_{avg} = \frac{1}{\sum_{i=1}^{n} w_i} \times \sum_{i=1}^{n} w_i I_i \qquad (12)$$

In (12), $n$ is the size of the sub-population and $I_i$ is the individual's self-improvement given by:

$$I_i = \left(P_{old_{actual i}} - P_{actual\ i}\right) \qquad (13)$$

In (13), $P_i$ is the individual's performance. The weight of an individual's vote, $w_i$ is calculated as follows:

$$w_i = -\log \frac{d_{i,g}}{\sum_{k=0}^{n} d_{k,g}} \times \frac{P_{actual\ i}}{\left(P_{actual\ g} - P_{actual\ i}\right) + \varepsilon} \qquad (14)$$

where $d_{i,g}$ is the Euclidean distance between the individual and their governor. $\sum_{k=0}^{n} d_{k,g}$ is the total distance between a governor and every individual in their sub-population. $P$ is the performance, and $\varepsilon$ is a very small positive number. In (14), the log scale is used to mitigate the impact of extreme changes in distance calculation.

In the next step, a linear regression is used to compute the adjusted performance of each governor based on their population average performance and votes. Given $I_{avg}$ and actual performance of each governor we calculate the predicted performance, $P_{predicted}$ as follows:

$$P_{predicted} = \beta_0 + \beta_1 \times I_{avg} + \varepsilon_i \qquad (15)$$

where $\varepsilon_i$ is the residual error whose distribution is $N(0, \sigma)$, and $b_0$ and $b_1$ are calculated as follows:

$$b_1 = \frac{\sum(I_{avg} - \overline{I_{avg}}) \times (P_{actual} - \overline{P_{actual}})}{\sum(I_{avg} - \overline{I_{avg}})^2} \qquad (16)$$

$$b_0 = \overline{P_{actual}} - b_1 \times \overline{I_{avg}} \qquad (17)$$

The adjusted performance, $P_{adjusted}$ is calculated using the following formula:

$$P_{adjusted} = P_{actual} + [\eta \times \Delta P] \qquad (18)$$

where

$$\Delta P = P_{actual} - P_{predicted} \qquad (19)$$

$$\eta = \frac{1}{n} \times \frac{s_j}{\sum_{i=1}^{n} s_i} \qquad (20)$$

In (20), *n* is the number of governors, and $s_j$ is the population size of the j[th] governor.

### 2.9 Genetic Operators: Recombination and Mutation

Finally, the genetic operators are used to generate offspring based on the initial population. To maintain genetic diversity, recombination and then mutation is applied to the existing solutions. Selection, crossover, and mutation are the three operators used in the Epistocracy algorithm. The crossover operator uses the tournament method to choose the best parents among the sampled candidates. Their chromosomes are split at randomly picked points between 1 and the chromosome size - 1. The chromosome of each of the two new offspring is made of the genes from the opposite side of the split point of each of the two parents.

A percentage of the new individuals are then mutated. Once a chromosome is selected to be mutated one of its genes is selected at random. This gene will be replaced with a random number between the upper and lower bounds. The chromosome is then validated to ensure all genes are within the bounds. If a gene is not within the bounds it will be set to the closest bound.

## 3 Experimental Results and Analysis

### 3.1 Evaluation of Epistocracy Algorithm Using Benchmark Functions

To test the performance of our proposed algorithm, we have used several multimodal benchmark functions (i.e. Eggholder, Rastrigin, Schaffer-4, CrossInTray, Griewank) with a large number of local optima from the global optimization literature. We have also used 5 state-of-the-art evolutionary algorithms to compare the consistency and reliability of the Epistocracy algorithm using these benchmark functions.

In order to make a fair comparison between Epistocracy and other state-of-the-art algorithms, we have selected a set of optimization problems, and tested each algorithm with a population size of 100, for 100 runs and 100 iterations in each run.

The results of comparison among Epistocracy, Genetic Algorithm, Evolutionary Strategies, Artificial Bee Colony, Cuckoo Search, and Particle Swarm Optimization on different functions are given in Table 1, where "Mean" indicates the average fitness obtained from 100 runs and "Std." is the standard deviation. "Min" and "Max" are the best and worst fitness values, found throughout 100 runs, respectively.

As shown in Table 1, the Epistocracy algorithm demonstrates higher reliability and consistency compared to other algorithms due to lower variation and dispersion in the outcome of the objective function.

**Table 1.** Comparison of the benchmark functions.

| Function | | Epistocracy | GA | ES | ABC | CSA | PSO |
|---|---|---|---|---|---|---|---|
| **Eggholder 2D** | Min | **-959.6407** | -959.6407 | -959.6407 | -959.6407 | -959.6407 | -959.6407 |
| | Max | -957.7592 | -894.4704 | -893.6453 | -951.0668 | -753.0372 | -786.5260 |
| | Mean | -959.5399 | -938.0387 | -946.7461 | -958.8248 | -917.5035 | -927.3717 |
| | Std. | 0.4198 | 16.6918 | 22.8390 | 1.9929 | 48.1905 | 49.8634 |
| **Rastrigin 5D** | Min | **7.1054E-15** | 0.7325 | 4.1369 | 1.5228E-05 | 0.8285 | 1.9903 |
| | Max | 7.8160E-14 | 7.0667 | 11.5452 | 0.0033 | 5.6601 | 14.9245 |
| | Mean | 2.7001E-14 | 3.2081 | 8.5620 | 0.0012 | 2.5568 | 6.4700 |
| | Std. | 1.5046E-14 | 1.6610 | 1.9900 | 0.0009 | 1.1924 | 3.1550 |
| **Schaffer-4 2D** | Min | **0.2926** | 0.2926 | 0.2926 | 0.2926 | 0.2926 | 0.2926 |
| | Max | 0.2926 | 0.3067 | 0.2956 | 0.2929 | 0.3037 | 0.2926 |
| | Mean | 0.2926 | 0.2965 | 0.2937 | 0.2927 | 0.2957 | 0.2926 |
| | Std. | 8.0929E-09 | 0.0035 | 0.0009 | 0.0001 | 0.0036 | 1.4528E-07 |
| **CrossInTray 2D** | Min | **-2.0626** | -2.0626 | -2.0626 | -2.0626 | -2.0626 | -2.0626 |
| | Max | -2.0626 | -2.0620 | -2.0623 | -2.0626 | -2.0626 | -2.0626 |
| | Mean | -2.0626 | -2.0625 | -2.0625 | -2.0626 | -2.0626 | -2.0626 |
| | Std. | 9.1125E-16 | 0.0002 | 7.5410E-05 | 8.7118E-09 | 4.2811E-08 | 9.1125E-16 |
| **Griewank 2D** | Min | **0.0000** | 0.0082 | 0.0133 | 2.6017E-07 | 5.1834E-09 | 0.0000 |
| | Max | 0.0101 | 0.1353 | 0.1556 | 0.0123 | 0.1320 | 0.0099 |
| | Mean | 0.0050 | 0.0561 | 0.0728 | 0.0036 | 0.0374 | 0.0035 |
| | Std. | 0.0037 | 0.0317 | 0.0399 | 0.0046 | 0.0370 | 0.0039 |
| **Griewank 5D** | Min | **4.9108E-11** | 0.2869 | 0.3320 | 9.1378E-06 | 0.0652 | 0.0246 |
| | Max | 0.0862 | 1.0702 | 1.4168 | 0.0639 | 0.5257 | 0.2463 |
| | Mean | 0.0398 | 0.7734 | 1.0838 | 0.0309 | 0.2215 | 0.1184 |
| | Std. | 0.0190 | 0.2264 | 0.2701 | 0.0188 | 0.1217 | 0.0609 |

As illustrated in Fig. 3, The absence of outliers and smaller standard deviation represented by a tinier boxplot are the most significant advantages of the Epistocracy algorithm. According to the test results of Rastrigin 5D, Epistocracy obtained the smallest standard deviation and produced better mean than other algorithms. This is a proof that the Epistocracy algorithm can effectively avoid being trapped in local minima in a complex, multimodal environment. This also shows that our algorithm is scalable and has a clear advantage over other evolutionary algorithms tested in this problem.

For Schaffer-4 2D, Epistocracy algorithm shows a higher reliability than other algorithms by producing results within a narrower range depicted in its corresponding boxplot. With CrossInTray 2D, the Epistocracy algorithm still is either doing better than other algorithms such as GA, and ES, or performing the same as PSO. However, overall, the Epistocracy algorithm has a better consistency and reliability than PSO, and similar algorithms. Among all other algorithms, for Griewank 2D, the Epistocracy algorithm has produced a narrower range of optimal solutions which is represented by its tiny boxplot. For Griewank 5D, the Epistocracy algorithm, again, shows a better result than other algorithms, and reconfirms the reliability and consistency of the algorithm working in different environments with different characteristics.

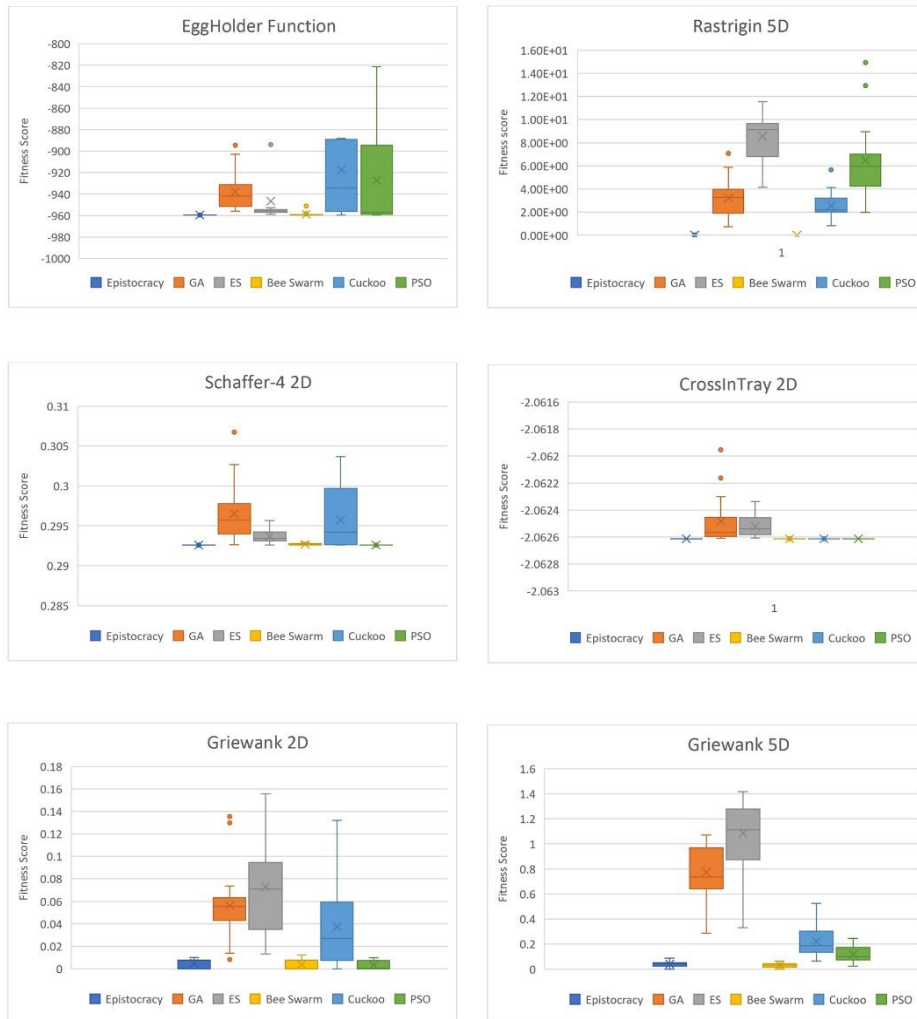

**Fig. 3.** Box and Whisker Plot of Fitness Scores for different benchmark functions.

These preliminary results also show that the Epistocracy algorithm is more reliable with functions that contain multiple minima. From the robustness aspect, the Epistocracy algorithm is more robust with respect to the existence of multiple minima. For large scale search space, the Epistocracy algorithm performs more efficiently than other algorithms. In terms of convergence, the Epistocracy algorithm showed a decent rate of convergence compared to other algorithms.

## 3.2 Evaluation of Epistocracy Algorithm Using the MNIST Dataset

To further evaluate the performance of our method, we tasked the Epistocracy algorithm to find the optimal set of hyper-parameters to build the best CNN model for "MNIST" handwritten digit recognition.

The MNIST dataset is a set of hand-written digit images ranging from 0-9. This dataset contains size-normalized, gray-scale examples of digits written by 500 writers that were centered in a 28x28 image and associated with a label from 10 classes. MNIST consists of a training set of 60,000 examples, and a test set of 10,000 examples and was constructed from NIST's (the US National Institute of Standards and Technology) Special Database 3 and Special Database 1 which contain binary images.

Each feature vector (row in the feature matrix) consists of 784 pixels (intensities) flattened from the original 28x28 pixels images. The end goal is to classify the handwritten digits based on a 28x28 black and white image. MNIST dataset is commonly used for training classification algorithms and benchmarking purpose.

**Optimization of Hyper-parameters.** The problem of finding the optimal value for hyper-parameter $\lambda$ is called hyper-parameter optimization. The main technique for finding such a value is to choose a value $\lambda_i$ from the trial set $\{\lambda_1, \lambda_2, \dots, \lambda_n\}$, to evaluate the response function $\Psi(\lambda)$ for each one, and return the $\lambda_i$ that worked the best as $\hat{\lambda}$. The optimization of hyper-parameters can be expressed as follow:

$$\hat{\lambda} \approx \underset{\lambda \in \Lambda}{\operatorname{argmin}} \, \mathbb{E}_{\sim \mathcal{G}_x} \left[ \mathcal{L}\left(x; \mathcal{A}_\lambda(X^{train})\right) \right] \equiv \underset{\lambda \in \Lambda}{\operatorname{argmin}} \, \Psi(\lambda) \tag{21}$$

In the above formula, $\lambda$ is the hyper-parameter that should be selected in a way that the generalization error (loss function) $\mathbb{E}_{\sim \mathcal{G}_x} \left[ \mathcal{L}\left(x; \mathcal{A}_\lambda(X^{train})\right) \right]$ minimized. $\mathcal{A}$ is the learning algorithm that maps the training dataset $X^{train}$ from a natural distribution $\mathcal{G}_x$ to the function f, $f = \mathcal{A}_\lambda(X^{train})$ The hyper-parameter optimization can be denoted as the minimization of the response function $\Psi(\lambda)$ over $\lambda \in \Lambda$ where $\Lambda$ is the search space.

**MNIST CNNs as a Proof of Concept.** Epistocracy as a multivariate optimization algorithm can be adapted for use in the automated discovery of CNN architectures, however, its effectiveness in doing so would be difficult to test. A regular multivariate optimization problem might have a known minimum or maximum while the accuracy of a CNN does not. In addition, the exact answer of most problems can be obtained through mathematical proof or exhaustive search. A full exhaustive search, however, is both time-consuming and computationally expensive, and there is no way to know what the best possible architecture of a model is.

The solution to this problem is to create a finite set of architectures and task Epistocracy with finding the best architecture in that set. For this purpose, a set of 480 unique models were generated, using all possible values of the hyper-parameters shown in Table 2. Every permutation of these hyper-parameters was used to create a distinct model. More hyper-parameters were not used since each model takes a considerable amount of time to train and test. Adding more options would make the amount of time needed to create all permutations of models unreasonable, and we must know the accuracy of all permutations in order to use MNIST as a proof of concept.

**Table 2.** Hyper-parameters and values used in an exhaustive search.

| Hyper-parameter | Values Used |
| --- | --- |
| Filter Number | 12, 16, 20, 24, 28, 32 |
| Filter Size | 3, 4, 5, 6, 7 |
| Neuron Size | 50, 100, 150, 200 |
| Dropout Rate | 0.1, 0.2, 0.3, 0.4 |

**Creating CNNs for MNIST.** The first step to making the 480 different architectures is to create every possible set of hyper-parameters. Each set is a unique set of hyper-parameters for a single model. Given a set of hyper-parameters, a 16-layer equivalent CNN model is created in Keras using Google's sample code to train an "MNIST" handwritten digit recognition model.

**Table 3.** Best combination of hyper-parameters.

| Hyper-parameter | Value |
| --- | --- |
| Filter Number | 28 |
| Filter Size | 6 |
| Dropout Rate | 3 |
| Neuron Size | 50 |

Once all 16 layers are created, the model is compiled with the "Adam" optimizer and "Categorical Cross-Entropy" loss function. The model is then trained with all 60,000 images in the MNIST training dataset. Each model was tested on the MNIST test set which consists of 10,000 images. After evaluating all 480 architectures the best combination of hyper-parameters was identified (see Table 3). The accuracy of this model was 99.51%.

Epistocracy was then used to search and find the same optimal set of hyper-parameters shown in Table 3. Epistocracy found the best answer 33% of the time. Of the 33% of runs which it found the best answer the answer was on average found around iteration 6. The mean accuracy of the best Governor was 99.48%. The configuration of the Epistocracy algorithm was as follows:

- Council rate: 10%
- Crossover Rate: 50%
- Mutation Rate: 20%
- Tournament size: 5
- Population size: 20

To evaluate the performance and robustness of the Epistocracy algorithm, we compared our proposed algorithm with two state-of-the-art algorithms: Particle Swarm Optimization and Genetic Algorithm. These algorithms were also tasked to find the best model's hyper-parameters similar to Epistocracy. The population, iterations, and number of runs the algorithm tested at are given below:

- Population size: 20
- Iterations: 100
- Runs: 100

**Table 4.** Comparison of GA, PSO, and Epistocracy algorithms using MNIST dataset.

|  | GA | PSO | EPISTOCRACY |
|---|---|---|---|
| MEAN ACCURACY OF TOP BEST PERFORMING GOVERNOR | **0.9948** | **0.9948** | **0.9948** |
| PERCENTAGE OF TIMES THE BEST ANSWER WAS FOUND | 26% | 28% | **33%** |
| AVERAGE NUMBER OF ITERATIONS BEFORE FINDING THE BEST ANSWER | 22.81 | 6.40 | **5.64** |

*Particle Swarm Optimization.* To compare how our algorithm performs against other algorithms, we used Particle Swarm Optimization (see Table 4). The configuration of PSO was the same as Epistocracy. The PSO mean accuracy was 99.48% and the best accuracy was found around the 7th iteration. The best accuracy was found 28 times out of 100 runs.

*Genetic Algorithm.* The other evolutionary algorithm tested was a genetic algorithm (see Table 4). The algorithm is a standard genetic algorithm using crossover and mutation. The implementation found the best answer 26 times out of 100 runs. The mean accuracy of the best Individual was 99.48% and the best accuracy was found around the 23[rd] iteration.

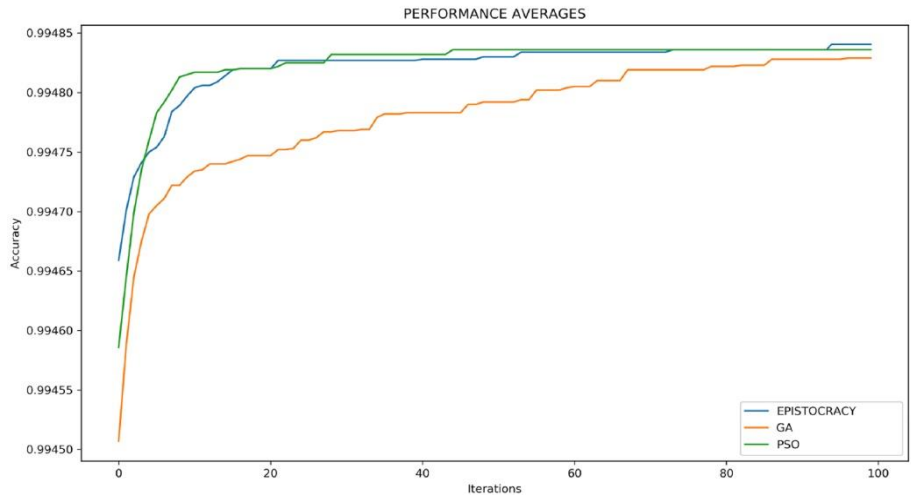

**Fig. 4.** Performance comparison of GA, PSO, and Epistocracy.

Fig. 4 shows that the Epistocracy algorithm initially has a higher accuracy than PSO and GA. After around 20 iterations, the Epistocracy algorithm asymptotically converges to the same fitness score of PSO, and eventually after about 92 iterations it defeats the PSO and shows higher accuracy. In this figure, the Epistocracy algorithm converges to the optimal solution faster than GA. However, even though PSO has a faster convergence rate of accuracy, it eventually fell behind Epistocracy. Overall, the Epistocracy algorithm shows better performance than the other algorithms.

## 4    Conclusion and Future Work

Evolutionary algorithms, in general, suffer from different types of problems such as premature convergence and stagnation which is closely related to the diversity of the population, curse of dimensionality and scalability, and a random, limited searching ability which usually occur in the absence of a guided change and due to unbalanced exploration-exploitation capacities.

This paper proposes a new multi-population evolutionary algorithm called Epistocracy based on socio-political evolution. In Epistocracy, there are two classes of population: governors and citizens. Citizens liberally follow governors to improve their performance through the exploration and exploitation of the search space. Governors, on the other hand, attempt to lead their population effectively to help the algorithm converge to the optimal solution in the early stages. Governors can be promoted or demoted based on their population performance and votes. Individuals with better performance have votes of greater weights.

The Epistocracy algorithm was tested using several benchmark functions. The experimental results show that the Epistocracy algorithm can achieve superior results compared to other evolutionary and swarm-intelligence algorithms. Our proposed method is less likely to be trapped in local optima compared to other methods such as

GA, PSO, ES, ABC, and CSA, and in some cases, can reach the optimal solution faster than existing algorithms. The Epistocracy algorithm uses the idea of rebels, dynamic resource management, gravitational force, and population variance to conduct an efficient explorative and exploitative search.

For future works, a number of research directions can be envisioned. First, the exploration-exploitation strategies can be enhanced to achieve a better convergence rate. Second, a multi-objective version of the algorithm can be implemented. Third, a more comprehensive test set with high dimensionality can be utilized, and the results compared with more evolutionary and swarm intelligence algorithms. Finally, the Epistocracy algorithm can be adapted for the discovery of optimal architectures of Convolutional Neural Networks and their hyper-parameters.